\documentclass{article}
\usepackage{spconf,amsmath,graphicx}
\usepackage{multirow}
\usepackage{booktabs,subcaption}
\usepackage{verbatim}
\usepackage{algorithm}
\usepackage{algorithmic}
\usepackage{caption}
\usepackage{color}
\usepackage{float}

\title{ADVKNN: ADVERSARIAL ATTACKS ON K-NEAREST NEIGHBOR CLASSIFIERS WITH APPROXIMATE GRADIENTS}
\name{Xiaodan Li, Yuefeng Chen, Yuan He, Hui Xue}
\address{Alibaba Group, China\\ \{fiona.lxd, yuefeng.chenyf, heyuan.hy, hui.xueh\}@alibaba-inc.com}
\begin{document}
\ninept
\maketitle
\begin{abstract}
Deep neural networks have been shown to be vulnerable to adversarial examples---maliciously crafted examples that can trigger the target model to misbehave by adding imperceptible perturbations.
Existing attack methods for k-nearest neighbor~(kNN) based algorithms either require large perturbations or are not applicable for large k.
To handle this problem, this paper proposes a new method called AdvKNN for evaluating the adversarial robustness of kNN-based models.
Firstly, we propose a deep kNN block to approximate the output of kNN methods, 
which is differentiable thus can provide gradients for attacks to cross the decision boundary with small distortions. 
Second, a new consistency learning for distribution instead of classification is proposed for the effectiveness in distribution based methods. Extensive experimental results indicate that the proposed method significantly outperforms state of the art in terms of attack success rate and the added perturbations.
\end{abstract}

\section{Introduction}
\label{sec:intro}
The security of deep learning models has gained tremendous attention, considering that they are the backbone techniques behind various applications, such as image recognition, translation, etc~\cite{ssd, maskrcnn,resnet,translation}.
Nonetheless, prior works mainly focus on higher accuracy, ignoring their robustness though adversaries can significantly affect the performance with small perturbations~\cite{moosavi2016deepfool, MIFGSM, PGD, brendel2017decision,Non-Parametric, TIattack, JSMA}, limiting the domains in which neural networks can be used, such as safepay with face recognition~\cite{parmar2014face} and self-driving~\cite{papernot2018deep}. 

To handle this problem, effective and generic defenses for deep models are proposed, such as adversarial training, data augmentation, distillation and etc~\cite{papernot2016distillation, sinha2019harnessing, cao2017mitigating, guo2017countering}. K-nearest neighbor~(kNN) based methods~\cite{papernot2018deep,sitawarin2019defending,dubey2019defense} are one of the toughest kind of defenses since they're non-parametric and can hinder gradients for the guidance of adversaries generation. 
For example, based on kNN, Deep k-nearest neighbors algorithm~(DkNN)~\cite{papernot2018deep} defenses the adversarial examples by ensembling kNN classifiers based on features extracted from each layer, which can effectively defense adversaries generated by FGSM~\cite{FGSM}, BIM~\cite{BIM}, CW~\cite{CW} attacks.

This work studies the problem of attacking kNN classifiers and evaluating their robustness. 
Previous attempts on attacking kNN models either apply gradient-based attacks to some continuous substitute models of NN~\cite{papernot2016transferability} or use some heuristics~\cite{sitawarin2019robustness}. For example, Papernot \emph{et al.}~\cite{papernot2016transferability} proposed to employ a differentiable substitue for attacking 1-NN model, which is not applicable for kNN model with large $k$; Sitawarin~\emph{et al.}~\cite{sitawarin2019robustness} proposed some heuristic methods to find adversarial targets and then use gradient-based model to find the optimal solutions to keep perturbations smallest. However, the distortions are still not small enough to be imperceptible.

In this paper, we propose a new adversarial attack called AdvKNN to attack kNN and DkNN classifiers with small distortions.
Firstly, we design a deep kNN block~(DkNNB) to approximate the output of kNN classifiers, which is differentiable so that it can guide the generation of adversarial examples with small distortion. 
To make the method more robust for DkNN, which summaries k nearest neighbors of each layer to get the final decision instead of the maximum probability like classical classifier does, 
we propose a new consistency learning~(CL) for probability distribution of k nearest neighbors instead of labels only.
Combined DkNNB and CL with simple attacks such as FGSM~\cite{FGSM}, BIM~\cite{BIM}, we find that both kNN and DkNN are vulnerable to adversarial examples with a small perturbation. Under $L_\infty$ norm~\cite{papernot2018deep}, our method manages to reduce the accuracy of DkNN on MNIST~\cite{MNIST} to only 5.71\% with mean $L_2$ distortion 1.4909, while Sitawarin~\emph{et al.} ~\cite{sitawarin2019robustness} got 17.44\% with distortion 3.476.

The main contributions of this paper are as follows:

1) We propose a deep kNN block to approximate the output probability distribution of k nearest neighbors, which is differentiable and thus can provide gradients to attack NN and DkNN models with small distortions. 

2) We propose a new consistency learning for distribution instead of classification, which makes our method more effective and robust for distribution based defenses. 

3) We evaluate our method on kNN and DkNN models, showing that the proposed AdvKNN outperforms prior attacks with higher attack success rate and smaller mean $L_2$ distortion. Besides, we show that the credibility scores from DkNN models are not effective for detecting our attacks.

\begin{figure}[tb] 
\centering 
\includegraphics[width=8cm]{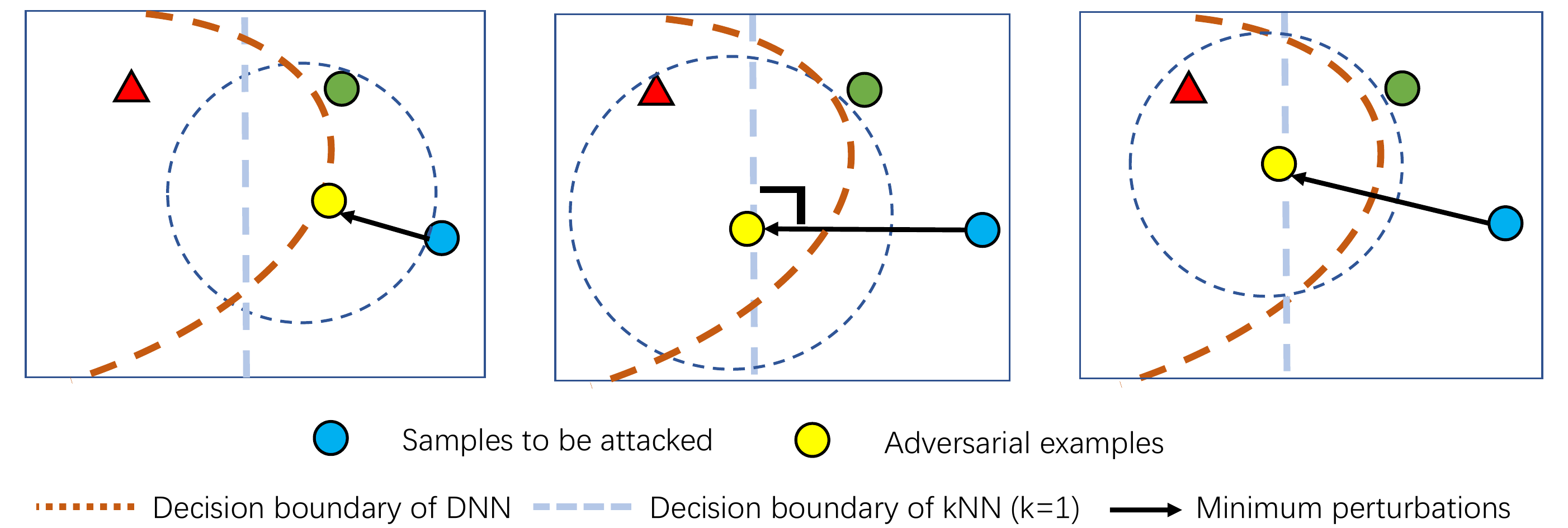} 
\caption{Illustration of minimum adversarial perturbation. From left to right, the yellow circles are the corresponding adversaries generated with minimum perturbations~(black lines with arrow) for DNN, kNN~($k=1$) and the method in~\cite{sitawarin2019robustness} respectively.} 
\label{Fig.attack} 
\end{figure}

\section{BACKGROUND AND RELATED WORK}
\label{sec:format}
In this paper, we focus on the adversarial examples in classification task based on deep neural networks~(DNN).
Adversarial example is a type of evasion attack against DNN at test time, that aims to find small perturbations to fool DNN models, which is defined as follows:
\begin{align}
\mathop{\arg\min}_\delta D(x, x+\delta) , \quad s.t. \, \, F(x+\delta) \neq y, x+\delta \in [0,1],
\end{align}
where $D$ is the distance metrics, $F$ is the DNN model, $y$ is the true label of $x$, and $\delta$ is the perturbation. 
Adversaries are generated by attack methods while defenses algorithms are designed to resist them.

\subsection{Classical Attacks}
\textbf{Fast Gradient Sign Method~(FGSM).}
FGSM~\cite{FGSM} is one of the most classical attack method. It's designed primarily for efficiency and optimized with $L_\infty$ distance metric, which controls the maximum absolute value of perturbation for one single pixel.
Given an image $x$, FGSM sets
\begin{align}
x^{\prime} = x + \epsilon \cdot \text{sign}(\bigtriangledown loss_{F,y}(x)),
\end{align}
where $\epsilon$ is chosen to be sufficiently small so as to be imperceptible.

\noindent\textbf{Basic Iterative Method~(BIM).}
BIM~\cite{BIM} is a refinement of FGSM. It takes multiple smaller steps $\alpha$ in the direction of the gradient sign and during each step the result is clipped by the same $\epsilon$ instead of taking one single step of size $\epsilon$. Specifically, $x_0^\prime = x$, while
\begin{align}
x_i = x_{i-1}^{\prime} + clip_{\epsilon}(\alpha \cdot \text{sign}(\bigtriangledown loss_{F,y}(x_{i-1}^{\prime}))), \quad & i \neq 0,
\end{align}
where $x_i^\prime$ is the generated adversarial example of input $x$ after step $i$.

\subsection{KNN-based Defenses}
\label{DkNN}
Fig.~\ref{Fig.attack} illustrates the smallest perturbation needed to attack DNN and kNN based classifiers.
As shown in Fig.~\ref{Fig.attack}, only a small perturbation is needed to cross the decision boundary when attacking normal DNN classifiers. To make models more robust, kNN-based methods are proposed. 
The kNN classifier is a popular non-parametric classifier that predicts the label of an input by finding its $k$ nearest neighbors in some distance metric such as Euclidean or cosine distance and taking a majority vote from the labels of neighbors~\cite{peterson2009k}.
The perturbation needed for kNN classifier is much larger than normal DNN classifiers, which makes attacks more difficult.

DkNN is a more robust kNN-based algorithm, which integrates predicted k nearest neighbors of all layers.
Denote $p_i^l(x)$ as probability predicted by kNN of class $i$ in layer $l$ of input $x$, then the final prediction of DkNN is 
\begin{align}
\text{DkNN}(x) = \mathop{\arg\max}_i \begin{matrix}\sum_{l=1}^L p_i^l(x)\end{matrix}, \quad s.t. \,\, i \in [1,2,\dots, C], 
\end{align}
where $C$ and $L$ are number of classes and layers respectively.
In addition to the final prediction, it proposes a metric called credibility to measure the consistency of k nearest neighbors in each layer.
The higher the adversary's credibility is, the model will treat it as clean sample with higher confidence. 
The credibility is computed by counting the number of k nearest neighbors of each layer from classes other than the majority, and this score is compared to the scores when classifying samples from a held-out calibration set. 
\begin{equation}
    cred(x) = \frac{1}{N}\sum_{n=1}^N J(\sum_{l=1}^L p_i^l(x) > \sum_{l=1}^L p_t^l(x_n)), \,\, s.t.\,\, i = \text{DkNN}(x),
\end{equation}
where $n \in [1,2,\dots,N]$. $x_n$ and $t$ are the $n$-th sample and corresponding true label in calibration set which has $N$ samples and
\begin{align}
J(y) = \begin{cases}
1, & y \text{ is True} \\
0, & y \text{ is False}
\end{cases}.
\end{align}

To evaluate the robustness of such kNN-based methods, Sitawarin ~\emph{et}\emph{ al.}~\cite{sitawarin2019robustness} mentioned that the minimum adversarial perturbation has to be on the straight line connecting $x$ and the center of training instances belonging to a different class, and once the target center is focused, they use gradient based method to find the smallest perturbation. However, as shown in Fig.~\ref{Fig.attack}, the optimal perturbation may not be on the lines connecting two points, but on the point who crosses the kNN decision boundary. Besides, 
the method proposed by Sitawarin~\emph{et al.}~\cite{sitawarin2019robustness} that finds adversaries in input space instead of feature space is not computationally efficient.

\begin{figure}[tb] 
\centering 
\includegraphics[width=8.5cm]{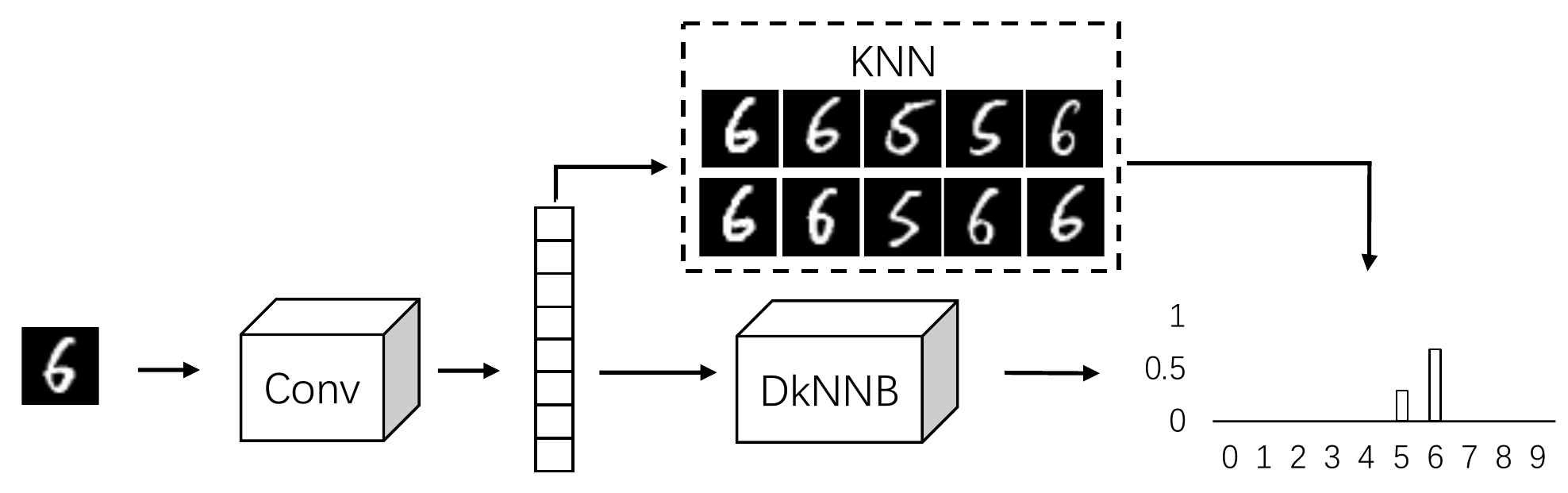} 
\caption{Illustraion of the proposed AdvKNN. The distribution of k nearest neighbors of inputs are used as the guidelines for optimizing DkNNB with CL.} 
\label{Fig.main2} 
\end{figure}


\section{ADVKNN: ADVERSARIAL ATTACK ON KNN}
This paper focuses on attacking kNN and DkNN classifiers. Our objective function can be denoted as:
\begin{multline}
\mathop{\arg\min}_\delta D(x, x+\delta), \quad \quad s.t.\, \,\, \text{kNN}(x+\delta) \neq \text{kNN}(x), \\
\text{DkNN}(x+\delta) \neq \text{DkNN}(x).     
\end{multline}

We detail our method with the specific introduction of DkNNB and consistency learning for distribution.

\noindent\textbf{Notation.}
Denote a predicted distribution of an input x as
\begin{gather}
f(x) = [f_1(x), f_2(x), \dots, f_C(x)], \quad  s.t. \, \,  \begin{matrix}\sum_{i=1}^C f_i(x) = 1\end{matrix},
\end{gather}
\begin{gather}
t = \mathop{\arg\max}_i f_i(x), \quad i \in [1,2,\dots, C],
\end{gather}
where $f_i(x)$ is the probability of $x$ belonging to class $i$ and $t$ is the predicted label of $x$.

\noindent\textbf{DkNNB.}
Let us start by precisely defining kNN. Assume that we are given a query item $x$, a database $G$ of candidate items $(g_i)_{i \in I}$ with indices $I = {1,\dots, M}$ for matching, and a distance metric $D(\cdot, \cdot)$ between pairs of items. Suppose that $x$ is not in the database,  $D$ yields a ranking of the database items according to the distance to the query. Let $\pi : I \rightarrow I$ be a permutation that sorts the database items by increasing distance to $x$:
\begin{gather}
\pi_x(i) < \pi_x(i^\prime) \Rightarrow D(x, g_i) \leq D(x, g_{i^\prime}), \quad \forall i, i^\prime \in I .
\end{gather}
The kNN of $x$ are the given by the set of the first $k$ items:
\begin{gather}
\text{kNN}(x, G) = \{g_i \mid \pi_x(i) \leq k\}.
\end{gather}
The kNN selection is deterministic but not differentiable. This effectively hinders to derive gradient to guide the generation of adversaries with small perturbation for methods such as FGSM, BIM and etc. We aim to approximate the prediction of kNN with neural networks, which can offer gradients for optimizing.

We focus on white-box threat model~\cite{brendel2017decision} for attacks on both kNN and DkNN, which means the attackers have access to the training set and all the parameters of target models. Since the training set as well as k are a kind of parameters of kNN classifiers as they are used during inference, we assume they are known by attackers as Sitawarin~\emph{et al.}~\cite{sitawarin2019robustness}.

To find the decision boundary of kNN classifiers, we propose a deep kNN block~(DkNNB). Specifically, the DkNNB is a small neural network which aims to approximate the output of k nearest neighbors of an input $x$. The illustration is presented in Fig.~\ref{Fig.main2}.

Suppose the distribution predicted by kNN and estimated distribution inferenced by DkNNB are $p(x)$ and $q(x)$ respectively. For a general kNN classification model, its prediction is a label $t$, and the corresponding distribution $p^o(x)$ is a one hot vector:
\begin{gather}
p_i^o(x) = \begin{cases}
1,& i = t \\
0,& others
\end{cases}.
\end{gather}

Then the loss for optimizing DkNNB towards kNN classifier is
\begin{equation}
\mathcal{L}_{CLS} = -\sum_{\forall x} p^o(x) \log q(x) = -\sum_{\forall x} \log q_t(x)    \label{crossentropy}.
\end{equation}

Then the derivative with respect to $x$ is:
\begin{equation}
\frac{\partial}{\partial x}\mathcal{L}_{CLS} = -\frac{\partial}{\partial x} \log q_t(x) \label{derive}.
\end{equation}

\noindent\textbf{Consistency learning for distribution.} 
As detailed in Section~\ref{DkNN}, DkNN summarizes probability distribution of outputs. However, as shown in Eq.~\eqref{derive}, learning with the classification loss only penalizes the true class, which is not optimal to approximate  kNN classification. Besides, the targets of kNN and DNN are correct classification labels. Learning the output labels of kNN will force the DkNNB to get the same parameters with the DNN classifier. To overcome these problems and further improve the attack performance, we propose to learn from the output distribution rather than  classification. We propose a new consistency learning~(CL) for distribution. Specifically, we define a new CL loss to guide the optimizing of DkNNB:
\begin{gather}
\mathcal{L}_{CL} = \sum_{\forall x} p(x)(\log p(x) - \log q(x)).
\end{gather}

Denote $\lambda$ as a hyperparameter, the final loss is:
\begin{gather}
\text{L} = \lambda \mathcal{L}_{CLS} + \mathcal{L}_{CL}.
\end{gather}


\begin{figure}[tb] 
\centering 
\includegraphics[width=8.5cm]{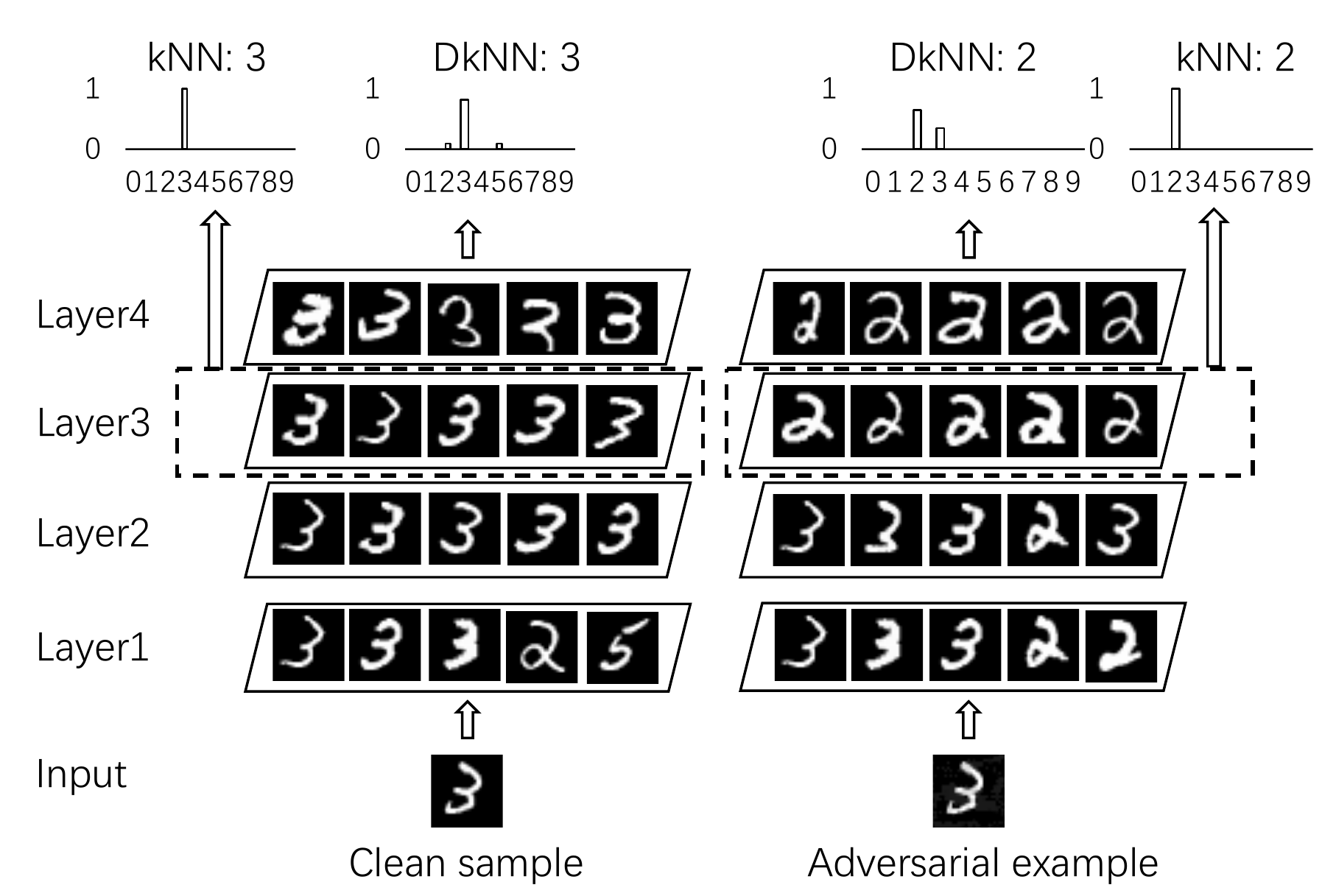} 
\caption{Comparison of k nearest neighbors of clean samples and adversarial examples. The left and right columns show the final decisions of kNN and DkNN as well as the neighbors in each layer of the clean sample and the corresponding adversarial example generated by the proposed AdvKNN. } 
\label{Fig.adv_samples} 
\end{figure}

\section{Experiments}
\label{sec:pagestyle}

\subsection{Datasets and Experimental Setting}
\noindent\textbf{Baselines \& Datasets.}
To demonstrate the effectiveness of the proposed AdvKNN, we evaluate our method on three datasets: MNIST~\cite{MNIST}, SVHN~\cite{SVHN}, and FashionMnist~\cite{fashion-mnist:}.
For each dataset, 750 samples (75 from each class) from the testing split are held out as the calibration set.
We reimplement DkNN from Papernot~\cite{papernot2018deep} with the same parameters as Sitawarin~\emph{et al.} did~\cite{sitawarin2019robustness} , including the base classifier network architecture and the value of $k=75$.
The proposed method can be applied to any gradient based attack algorithms. We choose FGSM, BIM as the baseline attacks which are optimized under $L_\infty$ norm~\cite{CW}. They are also used in DkNN for robustness evaluation~\cite{papernot2018deep}. Likewise, all the hyperparameters are set the same to the DkNN paper~\cite{papernot2018deep}.

The proposed DkNNB is implemented with a fully connected layer.
$\lambda$ is set to  0.3 for all the three datasets. If not mentioned,  kNN is conducted on the last convolution layer.
The classification accuracies of the backbone network~(DNN), kNN, DkNN as well as DkNNB are shown in Table~\ref{Table:acc}.

\subsection{Results and Discussion}
Fig.~\ref{Fig.adv_samples} shows a clean sample and its adversarial versions generated by our method along with their five nearest neighbors at each layer.
One the left column, all the majorities of the 5 neighbors in each layer belong to the same class as input. The final predictions of kNN and DkNN are correct. However, as shown on the right column, the majority of neighbors of the adversarial example are of the incorrect class. Both kNN and DkNN  are fooled successfully by adversaries with imperceptible distortions.

\begin{table}[!t]
\footnotesize
  \centering
  \caption{Classification Accuracy(\%) of DNN, kNN, DkNN and DkNNB.}
  \begin{tabular}{c|cccc}
  \toprule[1.2pt]
  {} & DNN & kNN  & DkNN & DkNNB \\
  \midrule[0.7pt]
  MNIST & 99.17 & 98.98  & 98.89 & 98.75 \\
  \midrule[0.7pt]
  SVHN & 89.93 & 88.37 & 90.35  & 88.32\\
  \midrule[0.7pt]
  FashionMnist & 91.22 & 89.87  & 90.28 & 89.23\\
  \bottomrule[1.2pt]
  \end{tabular}
  
 \label{Table:acc}

\end{table}

\noindent\textbf{Metrics.}
We employ  kNN accuracy, DkNN accuracy, mean $L_2$ distortion and mean credibility~\cite{papernot2018deep} to evaluate the effectiveness of our method~\cite{sitawarin2019robustness}. Lower kNN accuracy, DkNN accuracy, mean $L_2$ distortion and higher credibility mean better performance:
$\text{Attack success rate} = 1 - \text{Accuracy}.$

\begin{table*}[!htb]
\small
  \centering
  \caption{Evaluation results of the proposed DkNNB and Consistency Learning}
  \begin{tabular}{cc|ccc|ccc|ccc}
    \toprule[1.2pt]
    \multirow{2}{*}{Classifiers} &
    \multirow{2}{*}{Methods} &
    \multicolumn{3}{c|}{MNIST} & \multicolumn{3}{c|}{SVHN} &
    \multicolumn{3}{c}{FashionMnist}  
    \\
     & & Clean & FGSM & BIM & Clean & FGSM & BIM & Clean & FGSM & BIM \\
    \midrule[0.7pt]
    \multirow{3}{*}{kNN(ACC($\%$))~$\downarrow$} & Origin & 98.98 & 44.48 & 2.14 & 88.37 & 23.59 & 15.26 & 89.87 & 23.23 & \textbf{5.82} \\
    & DkNNB & 98.98 & 15.79 & 1.42 & 88.37 & 19.05 & 12.87 & 89.87 & 17.85 & 9.66  \\
    & DkNNB+CL & 98.98 & \textbf{15.28} & \textbf{1.03} & 88.37 & \textbf{16.37} & \textbf{9.28} & 89.87 & \textbf{11.87} & 7.54  \\
     \midrule[0.7pt]
    \multirow{3}{*}{DkNN(ACC($\%$))~$\downarrow$} & Origin & 98.89 & 54.90 & 16.80 & 90.35 & 28.60 & 17.90 & 90.28 & 31.65 & 16.28 \\
    & DkNNB & 98.89 & 28.57 & 6.49 & 90.35 & 22.89 & 13.98 & 90.28 & 30.38 & 14.87  \\
    & DkNNB+CL & 98.89 & \textbf{28.08} & \textbf{5.97} & 90.35 & \textbf{19.94} & \textbf{10.23} & 90.28 & \textbf{24.65} & \textbf{12.88}  \\
    \bottomrule[1.2pt]
  \end{tabular}
  \label{tab:attack_aware}
\end{table*}


\begin{figure}[tb]
\begin{subfigure}[b]{.48\linewidth}
    \centering
    \includegraphics[width=4.4cm]{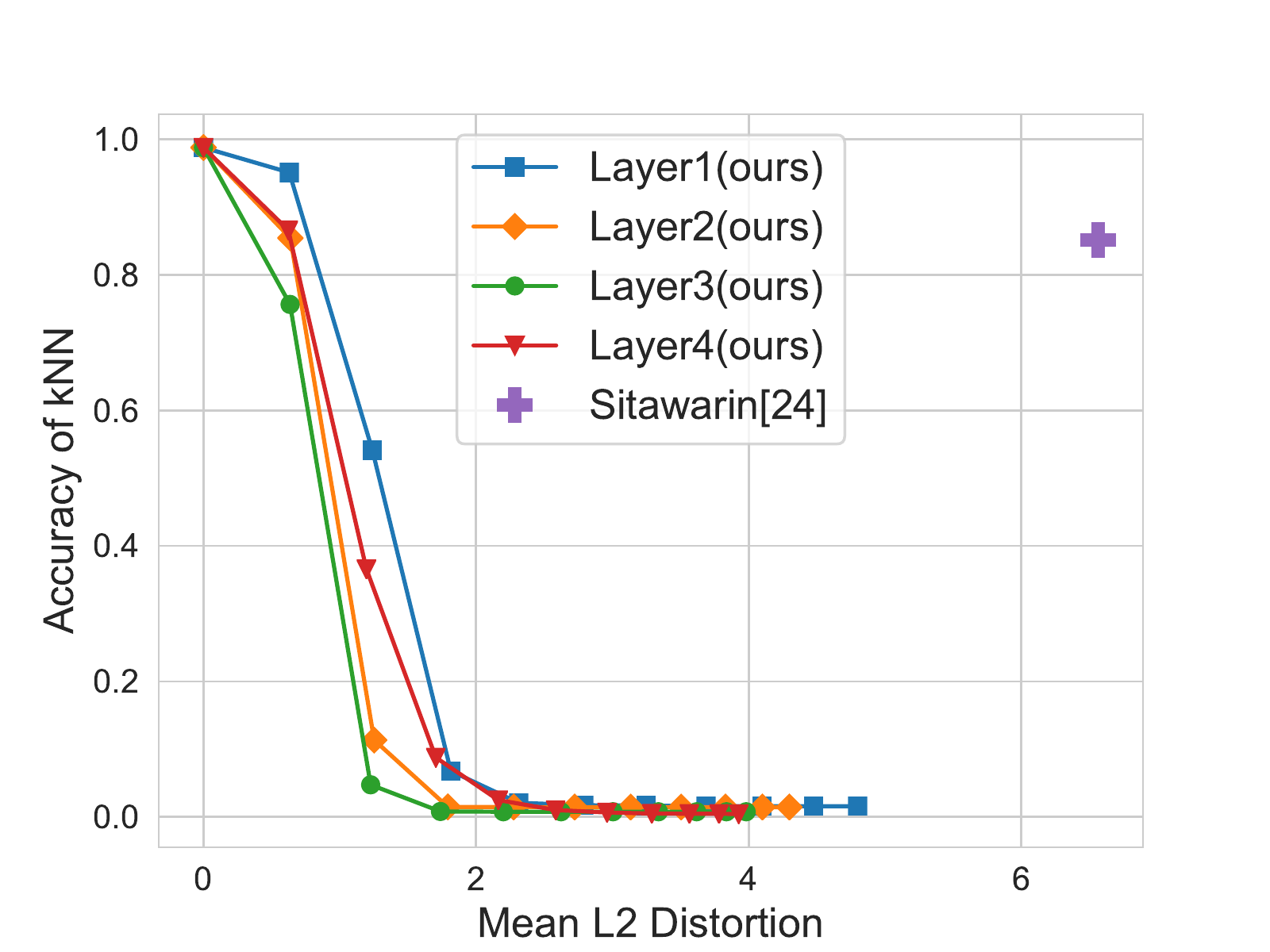}
    \caption{Accuracy of kNN}
    \label{Fig:kNN}
\end{subfigure}
\begin{subfigure}[b]{.48\linewidth}
    \centering
    \includegraphics[width=4.4cm]{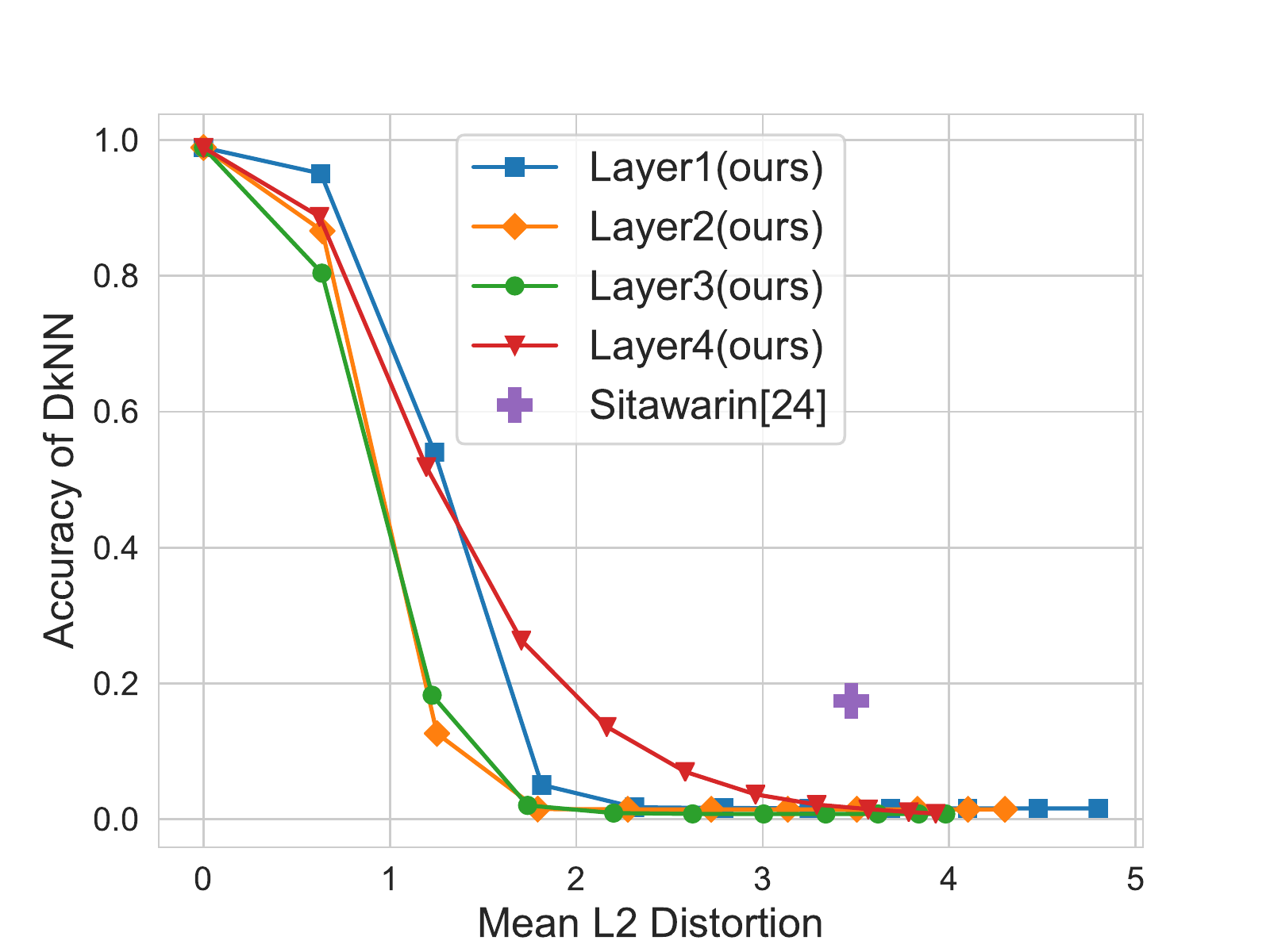}
    \caption{Accuracy of DkNN}
    \label{Fig:DkNN}
\end{subfigure}
\caption{Results of DkNNB and CL on different layers on MNIST.}
\label{Fig:performance}
\end{figure}

\begin{table}[t]
\footnotesize
  \centering
  \caption{Transferability of the generated adversaries on LeNet5.}
  \begin{tabular}{cc|ccc}
  \toprule[1.2pt]
  \multicolumn{2}{c|}{Accuracy($\%$) on LeNet5~$\downarrow$} & DNN & kNN & DkNN
  \\
  \midrule[0.7pt]
  \multirow{2}{*}{FGSM} & Origin & 68.42 & 75.06 & 81.41\\
  & DkNNB+CL & \textbf{56.06} & \textbf{64.97} & \textbf{75.70}\\
  \midrule[0.7pt]
  \multirow{2}{*}{BIM} & Origin & \textbf{47.46} & 59.72 & 72.61\\
  & DkNNB+CL & 47.55 & \textbf{56.37} & \textbf{70.04}\\
  \bottomrule[1.2pt]
  \end{tabular}
  
 \label{Table:transfer}

\end{table}

\noindent\textbf{Comparison with state-of-the-art methods.}
Table~\ref{tab:attack_aware} is the result of DkNNB added to the third layer. It shows the performances of attack methods before and after combining with the proposed method. With DkNNB, both FGSM and BIM attacks can degrade the classification accuracy of both kNN and DkNN by a large margin in most cases. 
After combined with CL, the performance is better.

\begin{figure}[tb]
\begin{minipage}[b]{.45\linewidth}
  \centering

  \includegraphics[width=4.4cm,height=3.4cm]{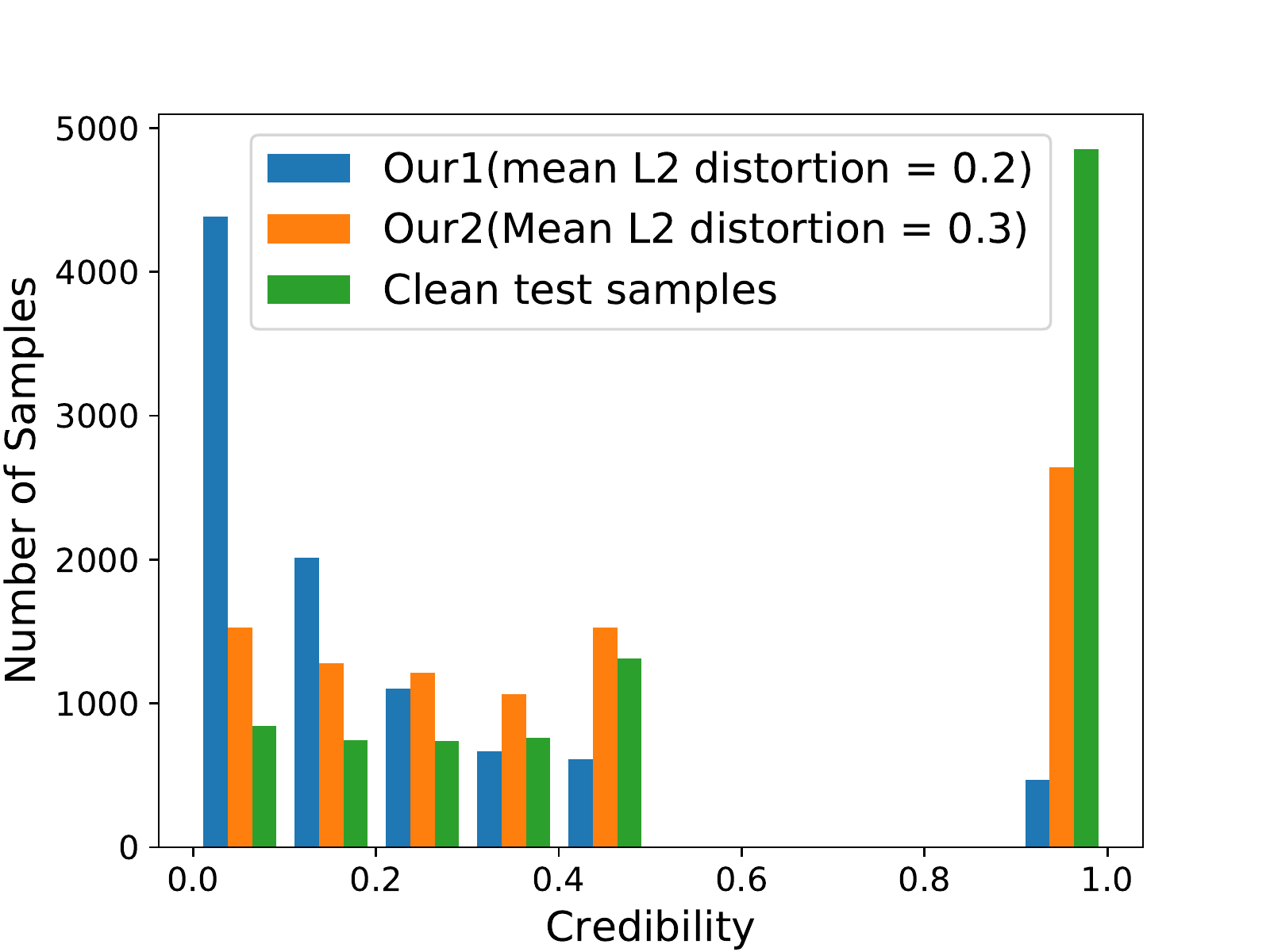} 
  \caption{Credibility of clean test samples and adversaries.}
  \label{Fig:Cred}
\end{minipage}
\hfill
\begin{minipage}[b]{0.45\linewidth}
  \centering
  \includegraphics[width=3.3cm]{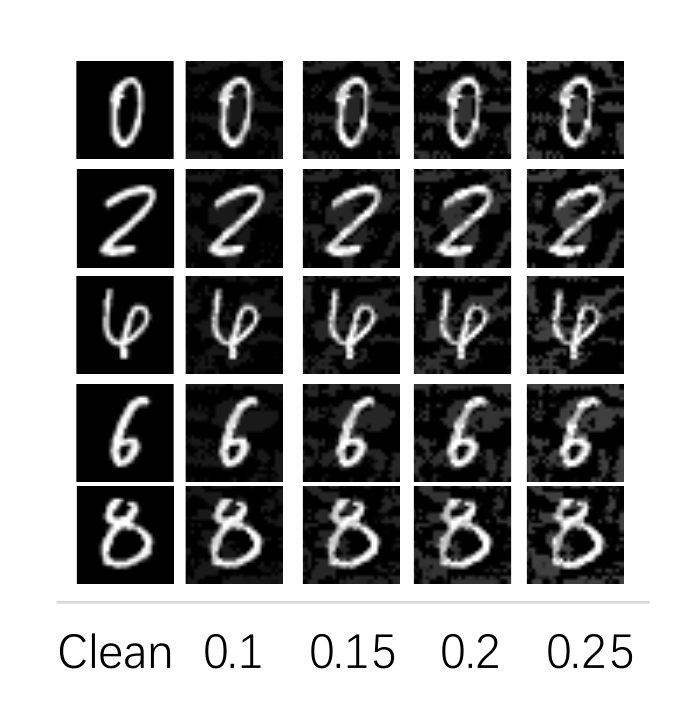}\caption{Generated adversaries under different $L_\infty$ norm}
  \label{Fig:advs}
\end{minipage}
\label{fig:ablation}
\end{figure}

Fig.~\ref{Fig:performance} shows the accuracies of kNN and DkNN changed with different mean $L_2$ distortions. The purple plus marker is the best attack result on DkNN with $L_\infty$ norm released by~\cite{sitawarin2019robustness}. It can be seen that our method can get the same accuracy drop with much lower $L_2$ distortion. 
In Fig.~\ref{Fig:Cred}, we show the distribution of credibility for the clean samples and adversarial examples generated by the proposed AdvKNN under different mean distortions. 
It can be observed that filtering out adversaries needs a lot of accuracy sacrifice of clean samples. With credibility score threshold 0.5, 71.42\% adversarial examples generated with mean $L_2$ distortion 0.3 can be detected, but 47.54\% clean samples will be filtered out too. Besides, mean credibility of methods proposed by Sitawarin~\cite{sitawarin2019robustness} is 0.1037 with mean $L_2$ distortion 3.476, while ours is 0.4608 with distortion  3.005, which indicates that our attack is more difficult to be detected. The generated adversarial examples are shown in Fig.~\ref{Fig:advs}.

For evaluating the transferability of generated adversaries to find if the generated adversaries can attack other models successfully, we further test the performance on LeNet5~\cite{lecun1998gradient} trained with clean samples of MNIST. 
Table~\ref{Table:transfer} shows the comparison of classification accuracy from different attack methods. It can be seen that adversaries generated with the proposed method performs better than FGSM and BIM, indicating the boost of adversarial examples transferability.

\begin{figure}[tb]
\begin{minipage}[b]{.5\linewidth}
  \centering
  \centerline{\includegraphics[width=4.6cm]{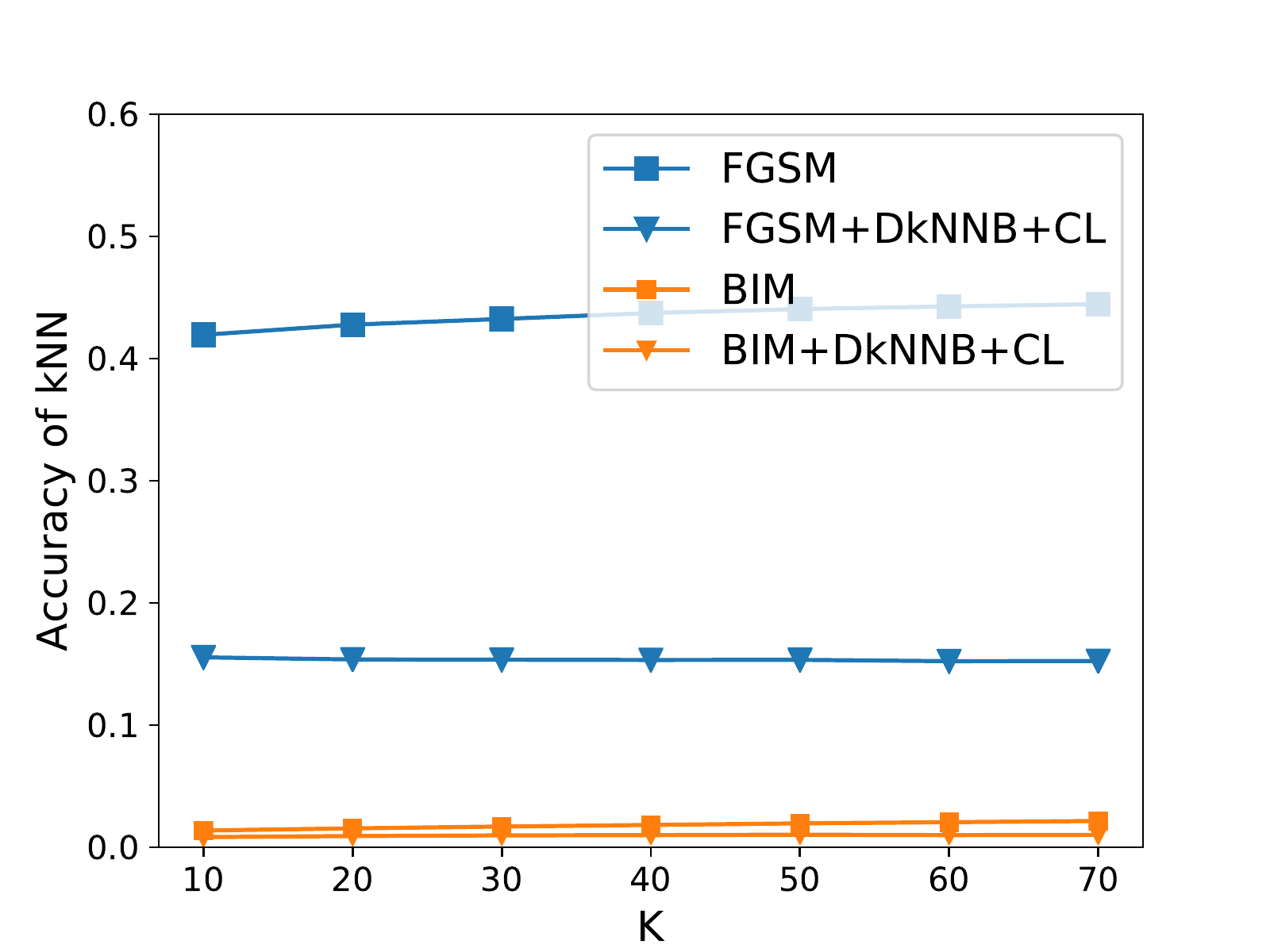}}
\end{minipage}
\hfill
\begin{minipage}[b]{0.5\linewidth}
  \centering
  \centerline{\includegraphics[width=4.6cm]{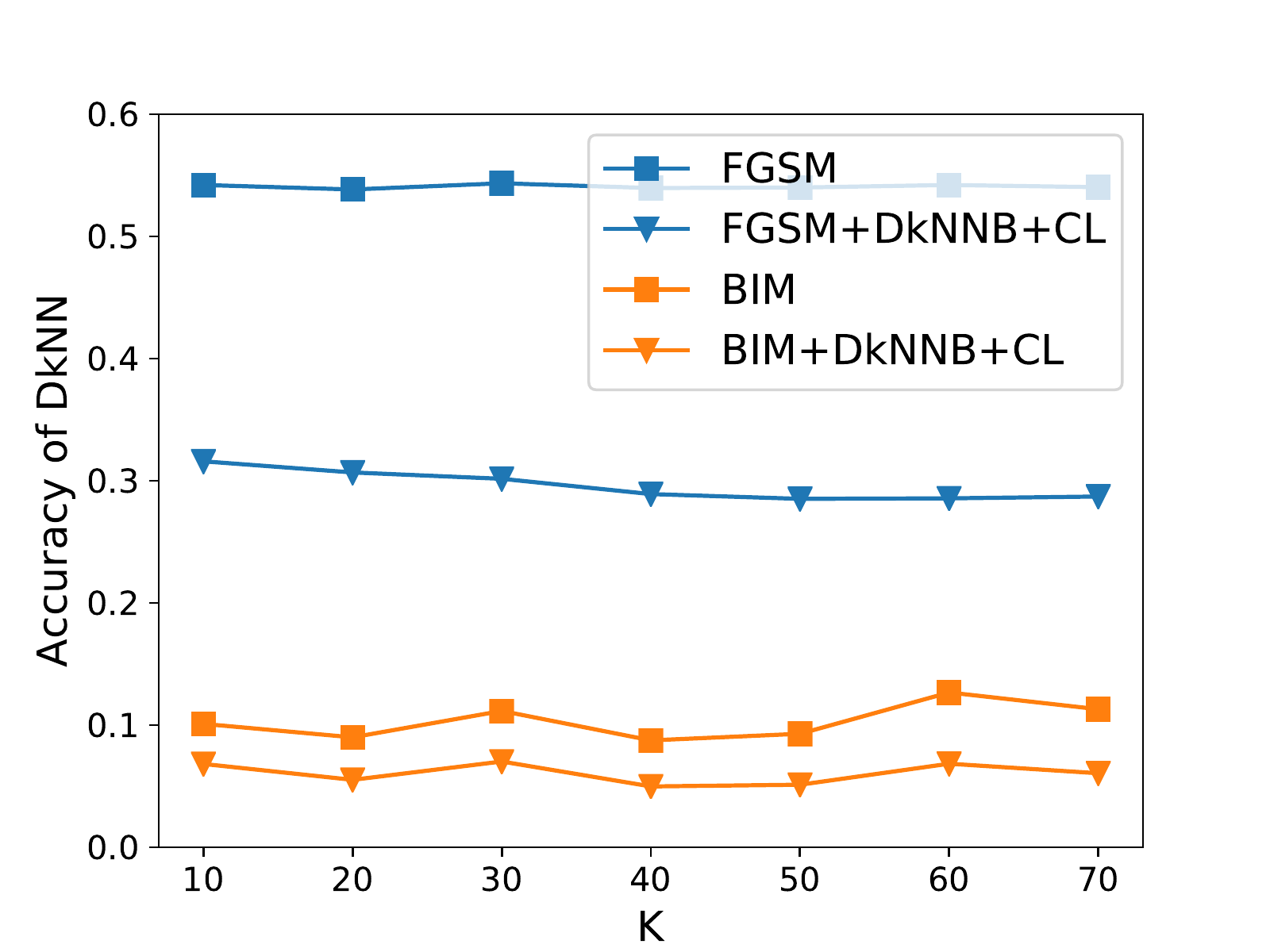}}
\end{minipage}
\caption{Comparison of performances with different $k$.}
\label{fig:k}
\end{figure}

\noindent\textbf{Ablation study.}
As shown in Fig.~\ref{Fig:performance}, the proposed DkNNB is always effective when equipped to each layer. Layer 3 performs the best in terms of kNN accuracy. This is reasonable because kNN is conducted on layer 3. 
It can be observed that the adversaries generated with DkNNB equipped to layer 2 also perform well, indicating the transferability between layers. Besides, we evaluate the attack performance under different $k$. As shown in Fig.~\ref{fig:k}, the proposed method always performs the best comparing with FGSM and BIM in terms of kNN and DkNN attack success rates.

\vspace{-0.1cm}

\section{Conclusion}
In this paper, we put forward a new method for attacking kNN and DkNN models to evaluate their robustness against adversarial examples, which can be easily combined with existing gradient based attacks. 
Specifically, we proposed a differentiable deep kNN block to approximate the distribution of k nearest neighbors, which can provide estimated gradients as the guidance of adversaries. Besides, we also proposed a new consistency learning for attack robustness on distribution based defense models.
We conducted extensive experiments and demonstrated the effectiveness of each part of the proposed
algorithm, as well as the superior overall performance.

\clearpage
\vfill\pagebreak


\bibliographystyle{IEEEbib}
\bibliography{strings,refs}

\end{document}